\documentclass[conference]{IEEEtran}
\IEEEoverridecommandlockouts
\usepackage{cite}
\usepackage{amsmath,amssymb,amsfonts}
\usepackage{booktabs}% http://ctan.org/pkg/booktabs
\newcommand{\tabitem}{~~\llap{\textbullet}~~}
\usepackage{algorithm}
\usepackage{algpseudocode}
\usepackage{graphicx}
\usepackage{textcomp}
\usepackage{xcolor}
\usepackage{bbm}
\usepackage{multirow}
\usepackage{hyperref}

\newcommand{\R}{\mathbb{R}}

\def\BibTeX{{\rm B\kern-.05em{\sc i\kern-.025em b}\kern-.08em
    T\kern-.1667em\lower.7ex\hbox{E}\kern-.125emX}}
    
\DeclareMathOperator*{\argmax}{arg\,max}
\begin{document}
%%%%%%%%%%%%%%%%%%%%%%%%%%%%%%%%%%%%%%%%%%%%%%%%%%%%%%%%%%%%%%%%%%%%%%
\title{AdverSAR: Adversarial Search and Rescue via Multi-Agent Reinforcement Learning
}
\author{\IEEEauthorblockN{Aowabin Rahman\textsuperscript{1}, Arnab Bhattacharya\textsuperscript{1}, Thiagarajan Ramachandran\textsuperscript{1}, Sayak Mukherjee\textsuperscript{1},  Himanshu Sharma\textsuperscript{1}, \\ Ted Fujimoto\textsuperscript{2}, Samrat Chatterjee\textsuperscript{3}\\
}
\IEEEauthorblockA{\textit{Optimization and Control Group}\textsuperscript{1}, \textit{Data Analytics Group}\textsuperscript{2}, \textit{Data Sciences and Machine Intelligence Group}\textsuperscript{3} \\
\textit{Pacific Northwest National Laboratory}\\
Richland, USA \\
% \{aowabin.rahman, arnab.bhattacharya, thiagarajan.ramachandran, sayak.mukherjee, samrat.chatterjee\}@pnnl.gov
}
% \IEEEauthorblockN{Arnab Bhattacharya}
% \and
}

\maketitle

\begin{abstract}
Search and Rescue (SAR) missions in remote environments often employ autonomous multi-robot systems that learn, plan, and execute a combination of local single-robot control actions, group primitives, and global mission-oriented coordination and collaboration. Often, SAR coordination strategies are manually designed by human experts who can remotely control the multi-robot system and enable semi-autonomous operations. However, in remote environments where connectivity is limited and human intervention is often not possible, decentralized collaboration strategies are needed for fully-autonomous operations. Nevertheless, decentralized coordination may be ineffective in adversarial environments due to sensor noise, actuation faults, or manipulation of inter-agent communication data. In this paper, we propose an algorithmic approach based on adversarial multi-agent reinforcement learning (MARL) that allows robots to efficiently coordinate their strategies in the presence of adversarial inter-agent communications. In our setup, the objective of the multi-robot team is to discover targets strategically in an obstacle-strewn geographical area by minimizing the average time needed to find  the targets. It is assumed that the robots have no prior knowledge of the target locations, and they can interact with only a subset of neighboring robots at any time. Based on the centralized training with decentralized execution (CTDE) paradigm in MARL, we utilize a hierarchical meta-learning framework to learn dynamic team-coordination modalities and discover emergent team behavior under complex cooperative-competitive scenarios. The effectiveness of our approach is demonstrated on a collection of prototype grid-world environments with different specifications of benign and adversarial agents, target locations, and agent rewards.

% can take significant advantage from supporting autonomous or teleoperated robots and multi-robot systems. These can aid in mapping and situational
% assessment, monitoring and surveillance, establishing communication networks, or searching for victims.
% This paper provides a review of multi-robot systems supporting SAR operations, with system-level considerations and focusing on the algorithmic perspectives for multi-robot coordination and perception.  
% We will demonstrate the effectiveness of the proposed approach in cooperative-competitive domains with sparse rewards where state-of-the-art methods fail and challenging multistage tasks necessitate dynamic modes of coordination.
\end{abstract}

%%%%%%%%%%%%%%%%%%%%%%%%%%%%%%%%%%%%%%%%%%%%%%%%%%%%%%%%%%%%%%
\section{Introduction}\label{sec:intro}
%%%%%%%%%%%%%%%%%%%%%%%%%%%%%%%%%%%%%%%%%%%%%%%%%%%%%%%%%%%%%%%
Multi-agent autonomous teams are becoming increasingly ubiquitous in a variety of Search and Rescue (SAR) missions that involve locating and rescuing targets in emergency situations. For example, military and first responders often deploy multi-robot teams to locate missing targets or survivors during disaster scenarios. In remote locations where communication with human operators may not be possible, a SAR team needs to search large areas of secluded geographical terrain in a fully autonomous and decentralized manner~\cite{queralta2020}. This can lead to a significant increase in time, effort, and operational costs when missing-target locations are not known \textit{a priori} with any degree of certainty. In such settings, inter-agent communications play a key role in enabling collaborative and coordinated decision-making~\cite{zhou2021multi}. This is because information shared by other robots (e.g. location of identified targets, terrain mapping) can help a robot reason about the current state of the world and narrow down it's search area rather than exploring independently using local information. Team performance in cooperative settings can significantly improve if agents systematically coordinate their exploration. 

However, multi-agent teams not only need to learn on how to cooperate, but also, on how to communicate effectively for cooperation. Even minor manipulations of messages shared within a team can significantly reduce the benefits of coordination~\cite{tu2021adversarial, behjat2021}. This destabilizing effect of perturbing communication data is further amplified in problems with sparse rewards~\cite{wang2020}. For example, in a SAR mission, the agents receive rewards only when some or all of the missing targets have been discovered and rescued (i.e., when a mission is completed successfully). Stealthy adversaries can strategically broadcast falsified sensor readings that prevents a SAR team from reaching or locating all of the targets despite active coordination between the agents~\cite{queralta2020}.

In this paper, we propose an adversarial multi-agent reinforcement learning (MARL) approach to learn efficient coordination and collaboration strategies to accomplish SAR mission objectives in the presence of adversarial communications. Our training approach is based on the well-known centralized training with decentralized execution (CTDE) paradigm~\cite{lowe2017multi} in MARL that uses full system information of all agents to generate rich features for learning effective decentralized agent-level policies. We have two major contributions in this paper. First, we developed a a general reward structure for MARL that allows effective  adversarial modeling and inference in sparse-reward settings. Second, we developed an adversarial multi-agent reinforcement learning algorithm that allows both cooperative and adversarial agents to be trained simultaneously. The effectiveness of our algorithmic approach is demonstrated for different environment and team specifications by reducing the overall target-exploration time of cooperative agents in adversarial presence.

%%%%%%%%%%%%%%%%%%%%%%%%%%%%%%%%%%%%%%%%%%%%%%%%%%%%%%%%%%%%%%%%
\section{Problem Description}\label{sec:problem}
%%%%%%%%%%%%%%%%%%%%%%%%%%%%%%%%%%%%%%%%%%%%%%%%%%%%%%%%%%%%%%%%

Our problem setup comprises of a team of cooperative agents tasked with covering a search area efficiently to locate missing assets (or targets). The locations of the missing assets are not known \textit{a priori}. Therefore, agents need to coordinate their respective search areas to avoid visiting the same locations multiple times. A mission is successful when all of the missing targets have been located within a pre-specified mission completion time. We make three assumptions in our model. First, it is assumed that all agents have homogeneous sensing capabilities. Second, agents cannot explicitly communicate with each other during mission execution. Finally, it is assumed that
target locations do not change over the search time of the SAR team.
%  %%%%%%%%%%%%%%%%%%%%%%%%%%%%%%%%%%%%%%%%%%%%%%
%  \subsection{Model Formulation}
% %%%%%%%%%%%%%%%%%%%%%%%%%%%%%%%%%%%%%%%%%%%%%%%
The problem is formulated as a Decentralized Partially Observable Markov Decision Process (Dec-POMDP) model~\cite{silver2015lecture} defined using a tuple $M=(S, A, O, P, R, Z, \gamma)$; here, $S$ denotes the finite set of states for all agents, $A$ is the joint set of actions, $O$ is the bounded set of observations, $P(s' | s, a) = P(s' = s_{t+1} | s = s_t, a = a_t)$ is the transition function, $R$ is the reward function, $Z(o_t | s_t, a_t)$ is the observation function and $\gamma$ is the discount factor. The goal of a general Dec-POMDP is to solve for a set of decentralized policies, $\pi = \{\pi^1, \pi^2, ... \pi^N\}$. The value function for a given policy $\pi$ starting from an initial state $s_0$ is expressed as \cite{omidshafiei2015decentralized}:
\begin{equation}
 V^\pi = \mathbb{E}_\pi \big( \sum_{t=0}^{\infty} R(s_t, a_t) | s_0 \big)
 \label{eq:POMDP_value}
\end{equation}
The solution to the Dec-POMDP is a policy $\pi^*$ that maximizes the value function $\pi^* = \argmax_\pi V^\pi$. Next, we propose an approach to model adversarial inference in a SAR context.
\section{State-of-the-art Model for SAR Exploration}
\label{sec:model_form}

\subsection{Summary of Current Approach}
\label{subsec:overall_approach}
%\color{blue} SM: section name needs to be edited \color{black}
For the naive case, i.e. SAR without the presence of adversaries, we leverage a two-tier MARL algorithm proposed in \cite{iqbal2019coordinated}. We would like to note that our key contributions in this paper concern modeling of adversarial presence and modifications to the algorithm by Iqbal and Sha \cite{iqbal2019coordinated} to account for adversarial presence -- which are detailed in section \ref{sec:adversary_model}. Nonetheless, we provide a brief overview of the the two-tier MARL approach in \cite{iqbal2019coordinated} as follows:
\begin{itemize}
    \item \textit{An upper-tier policy selection model:} A meta-policy model selects a single policy from a set of policies that all agents will adopt. Each policy from the set corresponds to a different collaboration strategy for exploration between the cooperative agents. We will discuss later on how each collaboration strategy is mathematically modeled as an intrinsic reward. The parameters of the upper-tier model, $\phi$, are learned via policy-gradient updates~\cite{iqbal2019coordinated}.
    
    \item \textit{A lower-tier soft actor-critic (SAC) model:} The lower-tier model optimizes individual agent policies (corresponding to each collaborative exploration strategy/novelty function). The model uses a central critic-network shared across all agents, and a policy model specific to each agent. For improved sample efficiency, the model is trained using off-policy learning methods, i.e.  all policies and value functions are learned using all data collected, regardless of which policy and exploration strategy were used to generate a given data sample.
    %The authors employ the soft version of the actor critic model, as it enables discourages convergence to sub-optimal policies (Harnooja et al. 2018).m 
\end{itemize}

\subsection{Intrinsic Rewards}
\label{subsec:intrinsic}

As mentioned in Section \ref{subsec:overall_approach}, each modality of team collaboration can be modeled as an intrinsic reward function. Since the SAR mission is a sparse-reward problem, i.e. the number of missing targets is small relative to the size of the domain, the intrinsic rewards serve a mechanism for training the agents to explore the domain efficiently. First, we define a novelty function $f_i$, which denotes how novel a given agent $i$ considers its own observations at a given location, based on past visits to that location. In a discrete environment domain (such as Gridworld), this can simply be the inverse of the state visits \cite{iqbal2019coordinated}. An intrinsic reward for agent $i$, denoted by $g_i$, is defined as a function that quantifies how novel all other agents consider the observations of agent $i$, and is expressed as
\begin{equation}
    g: [f_1(\mathbf{o_i}), f_2(\mathbf{o_i}), ...f_n(\mathbf{o_i})] \mapsto r_{intr}.
\label{eq:intr_reward}
\end{equation}
Here, $r_{intr}\in\R$ is a scalar value. For this paper, we consider three different exploration modalities, defined using three different intrinsic reward function $g_i$:
\begin{itemize}
    \item \textbf{Minimum} strategy rewards an agent when it explores a location no other agent has explored. The intrinsic reward for this case can be expressed as 
    \[g_i = \min_{j \in  \{1...N\}} f_j(o_i).\]
    \item \textbf{Covering} strategy rewards an agent when it explores a location that is more novel than the average agent and is defined as 
    \[g_i = f_i(o_i) \mathbbm{1} [f_i (o_i) > \mu(o_i)],\] where $\mu_i =\sum_j f_j(o_i)/N$.
    \item \textbf{Burrowing} strategy is the reverse of the covering strategy, as it rewards agents for exploring a location less novel than the average agent. This strategy leads to agents identifying the dead-ends quickly, which could lead to a reduced overall exploration time, and is expressed as 
    \[g_i = f_i(o_i) \mathbbm{1}[f_i (o_i) < \mu(o_i)],\]
    where $\mathbbm{1}(\cdot)$ denotes an indicator function.
\end{itemize}
Additional details on how to construct intrinsic reward functions is provided in \cite{iqbal2019coordinated}.
%%%%%%%%%%%%%%%%%%%%%%%%%%%%%%%%%%%%%%%%%%%%%%%%%%%%%%%%%%%%%
\subsection{Centralized Training with Decentralized Execution}
\label{subsec:CTDE}
%%%%%%%%%%%%%%%%%%%%%%%%%%%%%%%%%%%%%%%%%%%%%%%%%%%%%%%%%%%%%%%
Centralized Training with Decentralized Execution (CTDE) is an MARL training paradigm in which the agents share information during training (e.g. observations of other agents), but act on their own local observations during execution/evaluation \cite{lyu2021contrasting, lowe2017multi}. CTDE is useful in avoiding the non-stationarity issues that often arise in training multi-agent systems. Since a centralized critic in an actor-critic algorithm has access is used to observations of all agents, the agents are less likely to encounter dynamic changes in the environment -- which can lead to improved stability during training \cite{papoudakis2019dealing}. In the context of our problem,  a global state and an intrinsic reward function are used to train the policies of each agent, both of which require each agent having access to information of other agents. However, during execution, each agent acts based on its own policy, which only requires access to local observation specific to that agent.

% \begin{figure}
% \includegraphics[width=0.5\textwidth, trim=40 0 50 0,]{figures/fig_twotier_RL.pdf}
%   \caption{Schematic of two-tier reinforcement learning algorithm used for surveillance use-case without presence of adversaries, as proposed by Iqbal and Sha.}
% \label{fig:twotier_RL}
% \end{figure}
\section{Proposed Algorithmic Approach}
\label{sec:adversary_model}
\subsection{Adversarial Modeling}
\label{subsec:adv_model}
Here, we describe the SAR problem in the adversarial context. We introduce $N_{adv}$ adversarial agents (where $N_{adv} \in N$) in our domain, in addition to the cooperative agents already in the domain ($N_c \in N$). The overall goal of the adversary is to increase the time taken for the cooperative agents to find the missing assets. We make the following considerations in defining the adversary's reward structure.
\begin{itemize}
    \item The adversarial team has full knowledge of the location of the missing assets, and at each time-step, $t$, incurs a reward that discourages the cooperative agents to complete their mission. In our setup, we propose the following reward function for each adversarial agent:
\begin{equation}
    \hspace{-.4 cm} r_{adv}^{ext} = \alpha \sum_{i=1}^{N_c} \sum_{m=1}^{M_{nf}} |x_i - x_m| + |y_i - y_m|
\vspace{-0.5mm}
\label{eq:adversarial_reward}
\end{equation}
   \vspace{1mm}
    In Equation \ref{eq:adversarial_reward}, $(x_i, y_i)$ denotes the position of agent $i$ at a given time $t$, $(x_m, y_m)$ denotes the position of target $m$, $M_{nf}$ denotes the number of missing assets not discovered by the cooperative agents at time-step $t$, and $\alpha$ is a normalizing factor defined as $\alpha=\frac{K}{N_c(L+W)}$.
    Note that $N_c$ is the number of cooperative agents, $L$ and $W$ are the dimensions of a discrete environment, and $K$ is a hyper-parameter such that $0 < K \leq 1$. In this paper, we set $K=1$.
    \item The adversarial agents are rewarded for delaying the cooperative agents from completing the mission. Unlike the cooperative agents, the adversarial agents are not rewarded for locating the missing assets.
    \item If, at a given location in the search environment, an adversarial agent encounters a missing asset, it will spoof and falsify the location of the missing asset.
\end{itemize}

Table \ref{tab:reward_structure1} summarizes the reward structure for the cooperative and the adversarial agents, which will be referred to as the ``baseline" reward structure. Next, in section \ref{subsec:opt_coverage}, we describe a modified reward structure to enhance coordination under adversarial influence. 
\begin{table*}
\caption{Reward structure of cooperative agents vs. adversarial agents}
  \centering
  \begin{tabular}{lcc}
    \toprule
    \multicolumn{3}{c}{} \\[.5\normalbaselineskip]
    & Cooperative agents & Adversarial Agents \\
    \midrule 
    \textbf{External Rewards:} & \\
    \tabitem {Time penalty at each timestep}: &$-0.1$    & $+0.1$  \\
    \tabitem {Locating asset (not located  already by other assets)}: & $+10$    & $0$  \\
    \tabitem {Completing task (i.e. locating all assets)} & $+10$ & $0$ \\
    \tabitem Not completing task & -10 & 0 \\
    \tabitem Adversarial reward at each timestep, $t$  & N/A & Eq. \ref{eq:adversarial_reward} \\
    %\textbf{Intrinsic Rewards:} & \\
    \bottomrule
  \end{tabular}
  \label{tab:reward_structure1}
\end{table*}

%----Changes to the paper--------------------
\subsection{Reward Structure for Optimal Coverage}
\label{subsec:opt_coverage}
% - "Streamline/present" the algorithm as 
% - (a) Reward for baseline case, and (b) reward for optimal coverage, merge two sections.

We aim to extend the MARL algorithm for a general optimal coverage problem, and propose a ``modified" reward structure for both the cooperative and the adversarial teams. One key limitation of the existing model is that the primary extrinsic reward that the agents receive are dependent on target locations. This makes the trained two-tier MARL model difficult to generalize during evaluation for a new environment map and target locations. As such, we define a reward structure to encourage cooperative agents to explore the environment, irrespective of the location of the targets. We propose that at each time-step $t$, the cooperative agents are rewarded for exploring states that are novel, i.e. states that have not been visited by the cooperative team. Therefore, during training, we assign a secondary reward $r_{sec}$ signal defined as: 
\begin{equation}
    r_{sec, coop} = \sum_{i=1}^{N_c} \mathbbm{1} [v_{N_C} (x_i, y_i) = 1].
\label{r_sec1}
\end{equation}
In Equation \ref{r_sec1}, $v_{N_C}$ denotes the total number of visits by the cooperative agents at a given location ($x_i$, $y_i$) in the grid, i.e., $v_{N_C} = \sum_i^{N_c} v_i$ . Thus, Equation \ref{r_sec1} indicates that the cooperative team is rewarded when a given cooperative agent moves to a state not visited by the team previously. For the adversarial agents, the secondary reward signal is defined as
\begin{equation}
    r_{sec, adv} = \sum_{i=1}^{N_c} \mathbbm{1} [v_{N_C} (x_i, y_i) > v_{thresh})].
\label{r_sec2}
\end{equation}
Here, $v_{thresh}$ denotes a threshold value for the number of visits, beyond which a state visit is considered redundant (we set $v_{thresh} = 1$). The intuition behind the adversarial reward signal is that adversarial agents should encourage cooperative agents to visit redundant states that have been previously visited. The cumulative reward function is a weighted sum of the cooperative and adversarial team signals defined as 
\begin{equation}
    r = r_{sec} + \beta r_{intr},
\end{equation}
where $\beta$ is time-varying parameter defined as
\[
    \beta = 
\begin{cases}
  0.1,               & t\leq 0.4 T_{max},\\
  0.1\exp(-kt),      &\text{otherwise}.
\end{cases}
\]

Here, $T_{max}$ denotes the maximum number of time-steps per training episode. Note that the proposed reward structure for a general optimal-coverage problem does not require the terminal state to be dependent on location of targets. Thus, during training, we train the cooperative and the adversarial agents for a given map with no targets. During inference, however, we revert to the SAR setting requiring cooperative agents to locate their targets.

\begin{table*}
\caption{Reward structure for optimal coverage (as detailed in section \ref{subsec:opt_coverage}. }
  \centering
  \begin{tabular}{lcc}
    \toprule
    \multicolumn{3}{c}{} \\[.5\normalbaselineskip]
    & Cooperative agents & Adversarial Agents \\
    \midrule 
    \textbf{External Rewards:} & \\
    \tabitem {Time penalty at each timestep}: &$0$    & $0$  \\
    \tabitem {Locating asset (not located  already by other assets)}: & N/A    & N/A  \\
    \tabitem {Cooperative secondary reward at each timestep} & Eq. \ref{r_sec1} & N/A \\
    \tabitem Adversarial reward at each timestep, $t$  & N/A & Eq. \ref{r_sec2}  \\
    %\textbf{Intrinsic Rewards:} & \\

    \bottomrule
  \end{tabular}
  \label{tab:reward_structure}
\end{table*}

\subsection{Training Algorithm}

% The second contribution of the paper is to adapt the two-tier RL algorithm proposed by Iqbal and Sha \cite{iqbal2019coordinated} to incorporate adversarial presence. and propose mitigation strategies to account for adversarial impact. We will consider multiple steps to mitigate the impact of adversaries, as quantified in the increase in flow-time of the agents to complete the mission:
We leverage an adversarial MARL training procedure to train the cooperative agents in presence of adversarial agents. The adversarial actions inject continuous perturbations to the performance of the cooperative agents and, therefore, once the cooperative agents are trained in such scenarios, we can guarantee sufficient robustness to
malicious influences. We simultaneously train the adversary's action to minimize the reward obtained by the cooperative agents, thereby, making the cooperative agents aware of an intelligently trained adversary. To implement this idea, we follow \cite{pinto2017robust} to iteratively train the cooperative and adversarial agents. That is, we first train the cooperative agents keeping the adversary's policy fixed, followed by the cooperative agent's policy being fixed and adversarial policy being updated (as presented in algorithm \ref{alg:algo_with_adv}). We hypothesize that training the adversarial agents alternatively with the cooperative agents, as opposed to training both cooperative and adversarial agents together, will improve adversarial inference and effectiveness of the cooperative team's policies. 
% \begin{figure}
% \centering
% \includegraphics[width=0.5\textwidth, trim=40 0 50 0,]{figures/avg_reward_plot.pdf}
% \caption{Cooperative and adversarial rewards during training}
% \label{fig:training_rewards}
% \end{figure}
%We divide the total number of epochs into subgroups 

% \blue{Are we going to use the red color in algorithm or just for our own reference? I don't think we need to highlight that. Currently in the algorithm we have to bring those alternative training lines}
% \red{It would be great if Sayak and/or TI can add to the discussion here. @Sayak: Some pointers to other literature on alternate training and what motivated approach.}

% \blue{SM: Yes, will add it}

% \begin{itemize}
%     \item During training, we consider a subset of the agents to be adversarial. We have modified the two-tier algorithm to train the cooperative and adversarial agents, as shown in algorithm \ref{alg:algo_with_adv}.   As shown in algorithm \ref{alg:algo_with_adv}, we update the model parameters for critic and policy for the cooperative agents ($\phi_c$, and $\psi_c$), before updating the corresponding parameters for the adversaries. The updated training paradigm will allow both the team of cooperative and adversarial agents to maximize their own respective functions.
    
% \end{itemize}

\begin{algorithm}
\caption{Adversarial training for coordinated multi-agent exploration.}
\label{alg:algo_with_adv}
\begin{algorithmic}
\State {Initialize environment with $N$ agents: $N_{c}$ cooperative agents and $N_{adv}$ adversarial agents} 
%\Ensure $y = x^n$
\State Initialize replay buffer $D$ and $t_{update}\leftarrow0$ 
\State {nitialize choice for random asset locations}
\For {t = 1...total number of timesteps}
    \If {episode is completed}
        \For {j = 1...number of policies}
            \State UpdateSelector(R, h)
        \EndFor
    
    % \If {random asset locations == True}
    %     \State $\mathbf{s}, \mathbf{o} \leftarrow ResetEnv()$ with random asset locations
    % \Else
    %     \State $\mathbf{s}, \mathbf{o} \leftarrow ResetEnv()$
    % \EndIf
    % }
    \State Sample policy head: $h \sim \prod$
    \State Set $t_{ep} \leftarrow 0$ and reward $R\leftarrow 0$
    \EndIf
    \State Select actions for each agent $n$, $a_i \sim \pi_n^h(\cdot|{o_n})$, $n \in N$
    \State Send actions to environment to fetch $s', \mathbf{o'}, r_{coop}, r_{adv}$
    \State $R \leftarrow R + \gamma^{t_{ep}} r$ 
    \State Store tuple ($\mathbf{s}, \mathbf{o}, \mathbf{a}, \mathbf{s'}, \mathbf{o'}, r_{coop}$) in $D_1$ 
    \State Store tuple ($\mathbf{s}, \mathbf{o}, \mathbf{a}, \mathbf{s'}, \mathbf{o'}, r_{adv}$) in $D_2$ 
    \State Increment $t_{update} += 1$ and $t_{ep} += 1$
    \If {$t_{update}$ == steps per update}
        \For {j = 1..$N_{iter,c}$} 
            \State Sample minibatch, $B \sim D_1$
            \State Update critics using all coop agents, $n \in N_c$
            \State Update policies for all coop agents, $n \in N_c$
            \State Update targets for coop agents: $\bar{\phi_c}, \bar{\psi_c}$
        \EndFor
        \For {j = 1..$N_{iter,adv}$}
            \State Sample minibatch, $B \sim D_2$
            \State Update critics using all adv. agents, $n \in N_{adv}$
            \State Update policies for all adv. agents, $n \in N_{adv}$
            \State Update targets for adv. agents: $\bar{\phi_{adv}}, \bar{\psi_{adv}}$
               
        \EndFor
        \State Reinitialize $t_{update} = 0$
    \EndIf
\EndFor
\end{algorithmic}
\label{algo:training_algo}
\end{algorithm}

\section{Numerical Results}
\label{sec:results}
%%%%%%%%%%%%%%%%%%%%%%%%%%%%%%%%%%%%%%%%%%%%%%%%%%
\subsection{Experimental Setup}
\begin{figure}
    \includegraphics[width=0.50\textwidth, trim=40 0 50 0,]{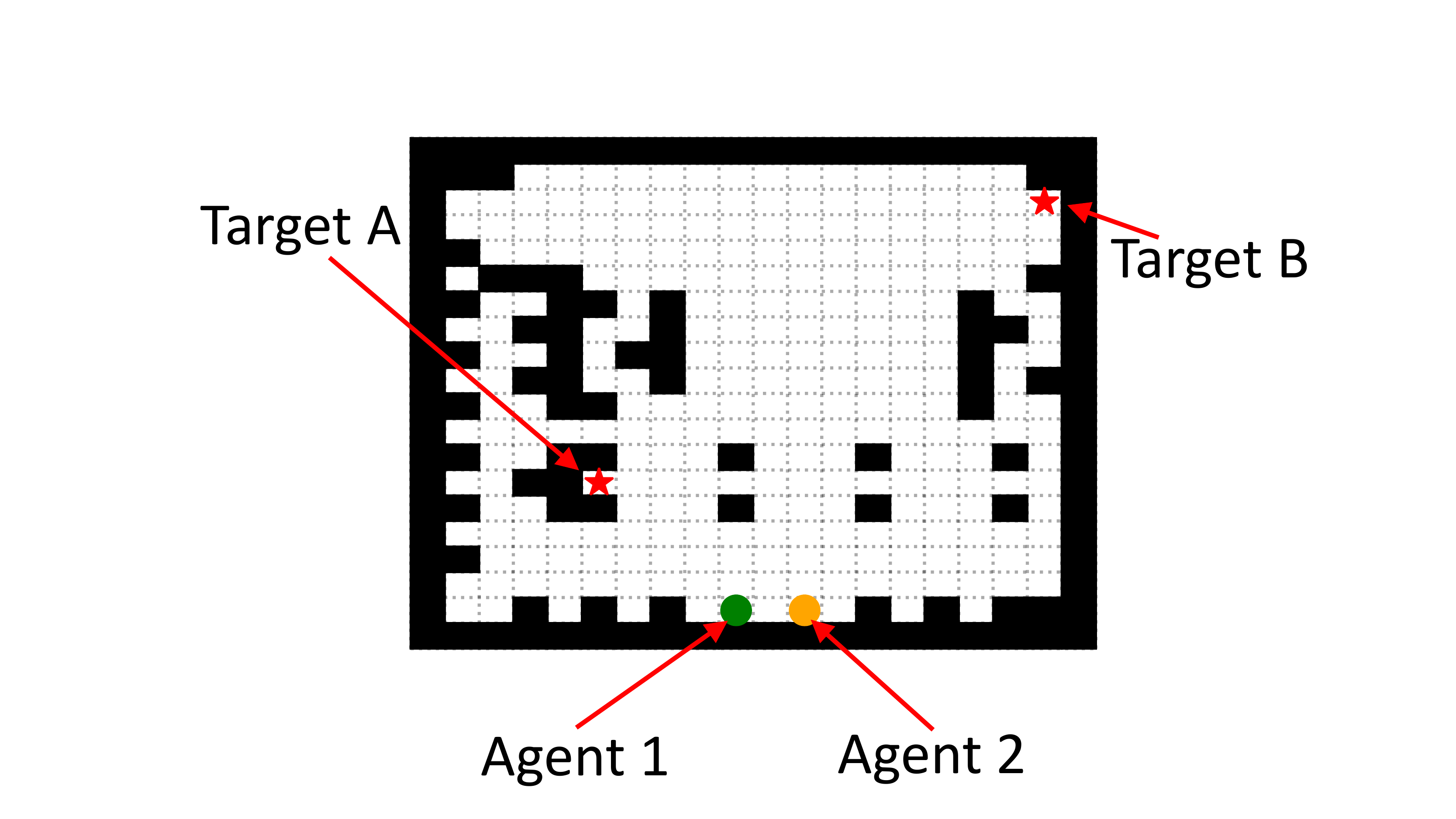}
      \caption{Schematic of Grid-world environment with two cooperative agents}
  \label{fig:schem_gridworld}
  \end{figure}

\begin{figure}
\vspace{-5mm}
    \includegraphics[width=0.50\textwidth, trim=40 0 60 0,]{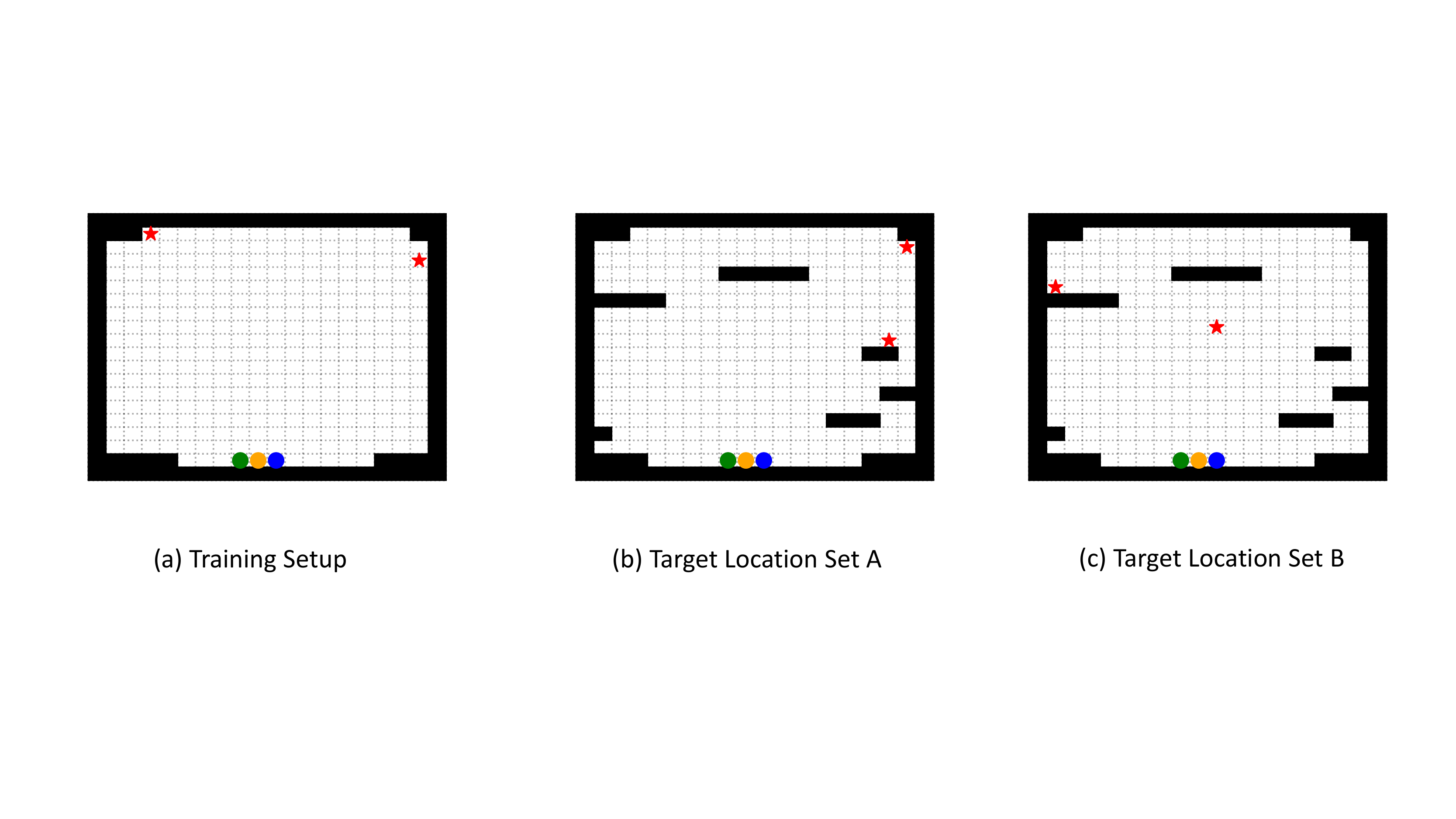}
    \vspace{-20mm}
      \caption{Grid-world maps used for training and inference: fig. (a) shows the map used for training (case III), whereas fig (b) and (c) show the inference maps with different sets of locations for the targets. Note that for case I, II and IV, we use a map identical to (a) for training, except that the target locations are not included.}
  \label{fig:maps}
  \end{figure}

%%%%%%%%%%%%%%%%%%%%%%%%%%%%%%%%%%%%%%%%%%%%%%%%%
We consider a 2D Grid-world domain (see Figure \ref{fig:schem_gridworld}) of size $20 \times 20$ as our simulation environment. Each cell within the domain represents the local position ($x_i, y_i$) of an agent $i$. At each time $t$, an agent can choose an action $a_i$ from a maximum of four actions; each action corresponding to taking a single step among four directions (going left, right, up or down). 
In our experiments, we aim to investigate: (a) how adversarial interference can derail performance of cooperative agents in a SAR mission; and (b) to what extent can the proposed training algorithm mitigate adversary's actions. We perform four case studies, as detailed in Table \ref{tab:case_studies}. We compare cases I vs. II to investigate objective (a) and cases III vs. IV to investigate objective (b). We evaluated the performance of the cooperative agents using the flow-time metric, which is defined as the number of time-steps needed for the cooperative team to locate the targets. For each case, we compute the mean flow-time by taking the average across 12 parallel instantiations of the environment. To investigate how well our models generalize, we evaluate the trained models on a different Grid-world map different to the one used during training, for two different target locations (see Figure \ref{fig:maps}).

\begin{table*}[htb!]
\caption{Case studies performed }
\centering
\begin{tabular}{|c|l|c|c|c|}
\hline
\textbf{Case Study} & \textbf{Description} & \textbf{\begin{tabular}[c]{@{}l@{}} Coop. agents\end{tabular}} & \textbf{\begin{tabular}[c]{@{}l@{}} Adversarial agents\end{tabular}} & \textbf{\begin{tabular}[c]{@{}l@{}}Reward \\ Structure\end{tabular}} \\ \hline
I & \begin{tabular}[c]{@{}l@{}}Algorithm with reward structure for optimal coverage.\\ SAC trained with two cooperative agents. \end{tabular} & 2 during training + inference & None & Modified \\ \hline
II & \begin{tabular}[c]{@{}l@{}}Training identical to case I. During inference, the \\ third cooperative agent is swapped for an adversarial agent.\end{tabular} & \begin{tabular}[c]{@{}l@{}}3 during training\\ 2 during inference\end{tabular} & 1 & Modified \\ \hline
III & \begin{tabular}[c]{@{}l@{}}Algorithm with reward structure specific to agents locating \\ missing assets. SAC trained with two cooperative and one \\ adversarial agent\end{tabular} & 2 during training + inference & 1 & Baseline \\ \hline
IV & \begin{tabular}[c]{@{}l@{}}Algorithm with reward structure for optimal coverage.\\ SAC trained with two cooperative and one adversarial agent.\end{tabular} & 2 during training + inference & 1 & Modified \\ \hline
\end{tabular}
\label{tab:case_studies}
\end{table*}

%%%%%%%%%%%%%%%%%%%%%%%%%%%%%%%%%%%%%%%%%%%%%%%%%%%%%%%%%%%%%%
\subsection{Adversarial Inference in a Cooperative SARSetting}\label{subsec:res_adv_model}
\begin{figure*}
   \centering
    \includegraphics[width=0.8\textwidth, trim=40 0 50 0,]{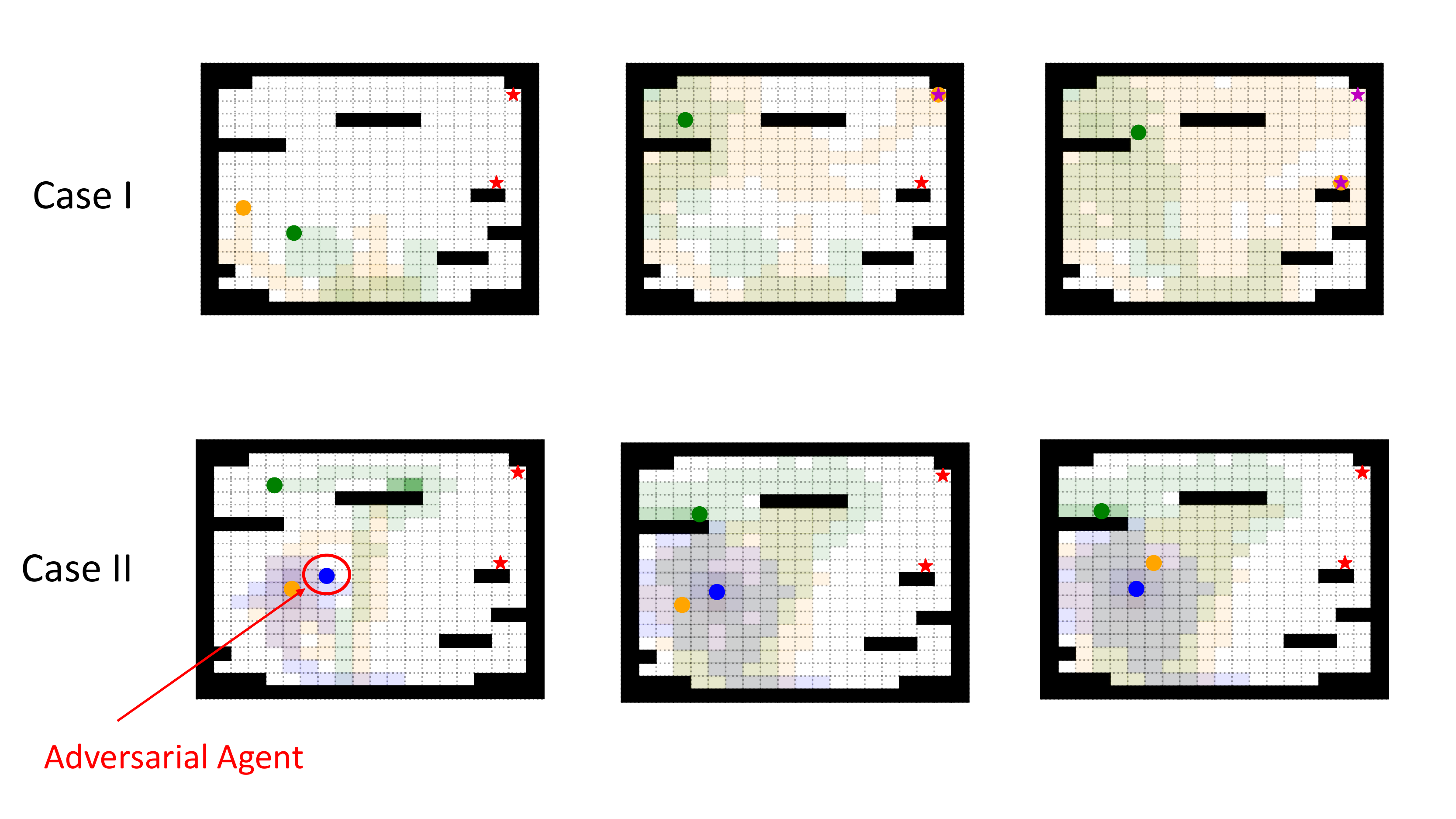}
      %\caption{}
   \caption{Illustration of agent trajectories during inference for a sample instantiation: (a), (b) and (c) correspond to preliminary, intermediate and terminal stages for cases I and II. For case II, the RL algorithm had been trained with three cooperative agents; but during inference, one of the coooperative agents was swapped for an adversarial agent (in blue). We observe that the adversarial agent prevents the cooperative agents from reaching the target locations. }
   \label{fig:case_i_vs_case_ii}
   \end{figure*}

%%%%%%%%%%%%%%%%%%%%%%%%%%%%%%%%%%%%%%%%%%%%%%%%%%%%%%%%%%%%
To illustrate the impact of adversarial interference, we consider the case where we first train the SAC model with three cooperative agents, but during inference, we replace one of the cooperative agents with an adversarial agent (case II). We compare this case against one where we train and infer with two cooperative agents (i.e. the number of cooperative agents remain identical across both cases). As reported in Table \ref{tab:inf_results}, the mean flow-times for case I is 1625, while for case II, the cooperative agent was unable to locate the target within 18,000 time-steps (which we set as the maximum number of time-steps during inference). Figure \ref{fig:case_i_vs_case_ii} illustrates the
comparison between Cases I and II during inference, for one of the 12 parallel instantiations. For both cases I and II, the agents adopt a ``burrowing'' strategy. While counter-intuitive, this strategy allows the agent to quickly identify dead ends in the map and then diverge to cover the domain. In Case I, the cooperative agents (in yellow and green) progressively cover more novel states to locate both of the missing assets, as seen from figure \ref{fig:case_i_vs_case_ii}(a)-(c).  However, the cooperative agents are thwarted, as they do not cover the sub-region where the missing assets are located. This is primarily because the right side of the domain was covered by a third cooperative agent during training, which, during inference, is swapped with an adversarial agent. Another reason for this could be that the adversary (blue agent) in Case II adopts a policy where it follows a trajectory close to one of the cooperative agents (in yellow). Since the observation of each agent includes whether another agent is present within three grid spaces, the adversarial presence in case II can ``spoof" the cooperative agent and prevent it from covering novel states. Thus, we observe that the presence of an adversary can significantly hinder a cooperative MARL task such as SAR. Therefore, it is critical to incorporate the presence of an adversary while training a multi-agent setup. 
% Later in this section (section \ref{subsec:res_adv_threat_mit}), we will observe how incorporating adversarial modeling as part of training can help the cooperative agents mitigate such adversarial threats. 
\begin{table}[htb!]
\begin{center}
\caption{Mean flow-time during inference for each of the four cases}
\begin{tabular}{|l|cc|}
\hline
\multirow{2}{*}{\textbf{Case}} & \multicolumn{2}{c|}{\textbf{Mean flow-time (timesteps)}} \\ \cline{2-3} 
 & \multicolumn{1}{c|}{\textbf{Map A}} & \textbf{Map B} \\ \hline
I & \multicolumn{1}{c|}{1625} & 1302 \\ \hline
II & \multicolumn{1}{c|}{\textgreater{}18000} & \textgreater{}18000 \\ \hline
III & \multicolumn{1}{c|}{\textgreater{}18000} & \textgreater{}18000 \\ \hline
IV & \multicolumn{1}{c|}{290} & 789 \\ \hline
\end{tabular}
\end{center}
\label{tab:inf_results}
\end{table}
%%%%%%%%%%%%%%%%%%%%%%%%%%%%%%%%%%%%%%%%%%%%%%%%%%%%%%%%%%%%%%%
\subsection{Adversarial Threat Mitigation using Proposed Algorithm}
%%%%%%%%%%%%%%%%%%%%%%%%%%%%%%%%%%%%%%%%%%%%%%%%%%%%%%%%%%%%%%
\label{subsec:res_adv_threat_mit}
We now investigate how effective the reward structures for optimal coverage (for the cooperative and adversarial agents) and the algorithm proposed in section \ref{sec:model_form} are in mitigating adversarial interference (case III), which we compare against a case in which we use the reward structure for cooperative structure as defined in \cite{iqbal2019coordinated}, and the adversarial reward as defined in Section \ref{eq:adversarial_reward}. We observe that in Case III, the cooperative agents head directly for the locations where the missing assets were located during training, since the episode rewards in this case were primarily dependent on finding the missing assets. As the location of the missing assets have changed from training to inference, this lack of generalizability leads to the cooperative agents failing to explore the domain efficiently and locate one of the missing assets within the allotted 18000 time-steps. In Case IV, where the rewards are not goal-conditioned (i.e. dependent on the location of the targets) and the impact of the adversary has been incorporated, the mean flow-time is 289.5. This supports our hypothesis that the proposed reward structure allows agents to generalize  novel states better.
   \begin{figure*}
   \centering
    \includegraphics[width=0.8\textwidth, trim=40 0 50 0,]{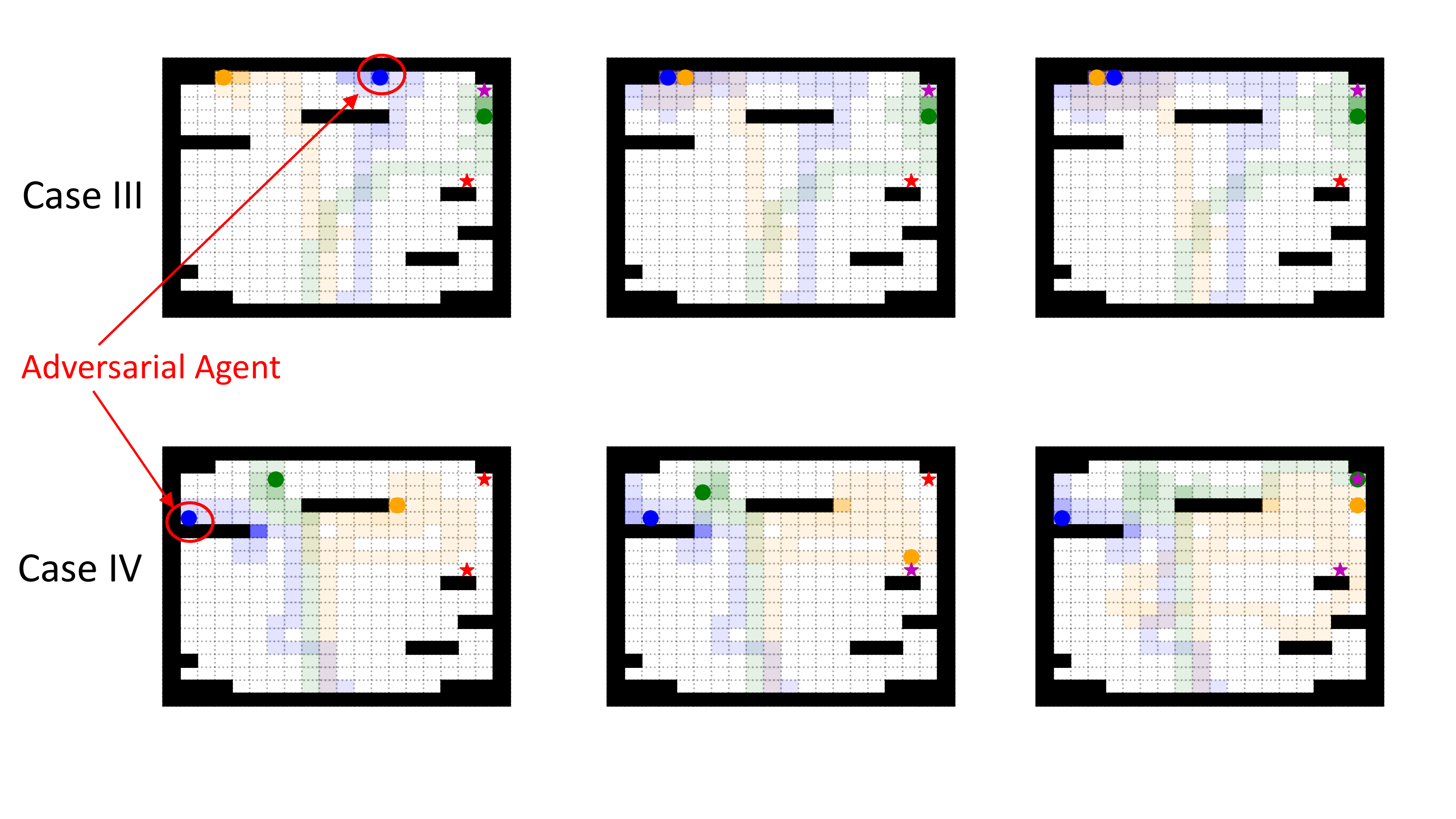}
      \caption{Illustration of agent trajectories during inference for a sample instantiation: (a), (b) and (c) correspond to preliminary, intermediate and terminal stages for cases III and IV. In case III, the reward structure and the terminal state during training are dependent on agents reaching the targets. During inference when the map and the target locations are altered, the cooperative agents failed to reach the targets within 18,000 time-steps. In case IV, with the modified reward structure, the agents explored the space more efficiently during inference and able to locate the targets.  }
   \label{fig:case_iii_vs_case_iv}
   \end{figure*}

\section{Conclusions and Future Work}
\label{sec:ongoing_work}
In this paper, we have shown that adversarial interference can significantly hinder the performance of a cooperative team in a SAR mission. We developed a reward structure that focuses on optimal coverage and not conditioned on target location, and an adversaril MARL training algorithm to learn cooperative team behavior that mitigates adversarial impact and improve generalizability in different SAR scenarios. As next steps, we seek to integrate our algorithm and demonstrate results in a higher fidelity 3D environment such as Microsoft AirSim. Moreover, we had assumed that the agents execute their decentralized policies based on their local observations, i.e. no communication between the agents were considered during execution. One possible research direction would be to investigate how adversarial interference could affect a cooperative team, where each agent is able to communicate with other agents within a given communication radius.

The code is open-source and hosted in github.com/aowabinr/SAR\_hst.

%%%%%%%%%%%%%%%%%%%%%%%%%%%%%%%%%%%%%%%%%%%%%%%%
\section*{Acknowledgment}
%%%%%%%%%%%%%%%%%%%%%%%%%%%%%%%%%%%%%%%%%%%%%%%%%
The research described here is part of the Mathematics for Artificial Reasoning in Science Initiative at Pacific Northwest National Laboratory (PNNL). It was conducted under the Laboratory Directed Research and Development Program at PNNL, a multiprogram national laboratory operated by Battelle for the U.S. Department of Energy under contract DE-AC05-76RL01830.

\bibliographystyle{IEEEtran}
\bibliography{references}

\end{document}